\pdfoutput=1

\documentclass[11pt,onecolumn]{article}

\usepackage{cvpr}
\usepackage{amsmath, amssymb, graphicx, xfrac, amsfonts, amsbsy, algorithm, color, subfig, url, multirow, hyperref}
\usepackage[noend]{algpseudocode}
\usepackage[section]{placeins}
\usepackage{float}
\cvprfinalcopy

\def \y {\mathbf{y}}
\def \x {\mathbf{x}}
\def \z {\mathbf{z}}
\def \u {\mathbf{u}}

\def \Th {\pmb{\Theta}}

\def \Rspace { \mathbb{R}}

\def\argmax{\mathop{\mathrm{arg\,max}}}

\begin{document}




\title{SpectroscopyNet: Learning to pre-process Spectroscopy Signals without clean data}


\author{ Juan Castorena and Diane Oyen}

\maketitle

\begin{abstract}
{\normalfont
	In this work we propose a deep learning approach to clean spectroscopy signals using only uncleaned data. Cleaning signals from spectroscopy instrument noise is challenging as noise exhibits an unknown, non-zero mean, multivariate distributions. Our framework is a siamese neural net that learns identifiable disentanglement of the signal and noise components under a stationarity assumption. The disentangled representations satisfy reconstruction fidelity, reduce consistencies with measurements of unrelated targets and imposes relaxed-orthogonality constraints between the signal and noise representations. Evaluations on a laser induced breakdown spectroscopy (LIBS) dataset from the ChemCam instrument onboard the Martian Curiosity rover show a superior performance in cleaning LIBS measurements compared to the standard feature engineered approaches being used by the ChemCam team.
}
\end{abstract}


\section{Introduction}
\label{Sec:introduction} 

Convolutional neural net (CNN) based denoisers have shown excelling performances over feature engineered approaches in both signal/image problem domains. The work in \cite{zhang2017beyond} for example, trained a discriminative CNN under supervision capable of removing additive white Gaussian noise very efficiently. Since then, extending such capabilities in the unsupervised setting have been the focus of study. Noise2Noise \cite{lehtinen2018noise2noise} and \cite{laine2019high} for example operates under self-supervised principles by imposing an $\ell_2$-norm reconstruction error over unpaired images with noise distributions drawn under a stationarity assumption. The deep model implictly learns the noise distribution from the unpaired images and explicitly discriminates the noise component from the image. One limitation however, is that it operates under the assumption that noise is drawn from a centered (i.e., zero mean) distribution, not attainable in many real-world applications.

One necesary condition for noise-to-signal disentanglement of the true generative factors is that of identifiability. This condition implies that no discriminative/generative model no matter how shallow or deep and inspite of learning in the limit of infinite data can disentangle the true generative factors of interest unless additional inductive biases are imposed \cite{hyvarinen1999nonlinear,locatello2020weakly}.
The works of \cite{zhang2017beyond} and  \cite{lehtinen2018noise2noise, laine2019high} for example, show highly effective discriminative learning methods in the supervised and self-supervised settings, respectively, all assuming a zero mean Gaussian noise distribution. Departing, however, from these assumptions does not offer the same performance guarantees and tends to produce biased disentanglements.

In this work, we utilize a simplified model of the data generative process by means of directed acyclic graphs (DAG)'s which are typically used in causal inference \cite{pearl1995causal}. DAG's provide a clear picture of the stated assumptions and enables graphical analysis to characterize the conditions under which identification is possible in more general non-zero mean and unknown but stationary distributions. Such identification requires only pairwise weak-labels to differentiate whether randomly chosen pairs of LIBS measurements were collected under target variability (i.e. that come from different targets) or if they are repeated (come from the same target). Our experimentation with a laser induced breakdown spectroscopy (LIBS) instrument shows that the proposed method is capable of effectively disentangling unknown non-zero mean distribution noise with superior performance compared to the standard methods used by LIBS expert practitioners.


\section{Approach}
\label{Sec:approach} 
\subsection{Problem formulation}
The data generation model we consider embeds the effects of the target of interest to the unmeasured signal component $\u_{x} \in \Rspace^N$ while those of noise to the unmeasured component $\u_n \in \Rspace^N$. 
Measurements $\y \in \Rspace^N$ collected  from such a generative model are assumed to be additive and given by:
\begin{equation} \label{model}
\y = \u_{\x} + \u_{n}
\end{equation}
where both $\u_{\x},\u_n$ are drawn from unknown non-zero mean, stationary multivariate distributions $p(U_X)$, $p(U_N)$. The problem we focus consists on learning neural representations $\z_{\x} \in \Rspace^N$ of the true signal $\u_{\x}$. In other words, disentanglement of the signal from the noise sources.

\begin{figure}[h]
	\centering 
	\includegraphics[width=0.2\linewidth]{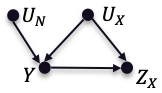}	
	\vspace{-0.75em}
	\caption{Directed Acyclic Graph of Generative Model.}
	\label{fig:DAG} 
\end{figure}

The generative model of Eq.\eqref{model} is generally not-identifiable unless a suitable parametrization of $\u_n$ is assumed; Fig. \ref{fig:DAG} illustrates the main issue. Conditioning on the measurements $Y$ opens up the free flow of information between $U_N \rightarrow Y \leftarrow U_X \rightarrow Z_X $ which biases the estimates of the signal representation $Z_X$ with unmeasured noise $U_N$.  Controlling for $U_X$ or $U_N$ would render the model identifiable, however, this is infeasible as both factors are unobserved. 

We make use of the assumption that noise $U_N$ is stationary and leverage data variability in $U_X$ in the limit of a very large sample for identification. We construct a proxy for $U_N$ referred to as $Z_N$ and use it as $p(U_N) = \sum_{Z_N} p(U_N | Z_N) p(Z_N)$.  This proxy is learned by a deep model trained to extract similarities between pairs of measurements that have target variability  (i.e., that come from different targets). For this purpose, we make use of the InfoNCE training objective \cite{oord2018representation} aiming at learning representations encouraging the mutual information (MI) \cite{oord2018representation} between distributions as
\begin{equation} \label{MI}
	\tilde{\Th_2} =  \argmax_{\Th_2}  \left\{ \textbf{MI} \left ( p( Z_n|Y) || p( Z'_n|Y') \right) \right \}
\end{equation}
Eq.\eqref{MI} provides the mechanism for isolating the effects of $U_N$ through the proxy $Z_N$ in the limit of very large sample size.

\subsection{Deep Model}
\begin{figure}[h]
	\centering 
	\includegraphics[width=0.5\linewidth]{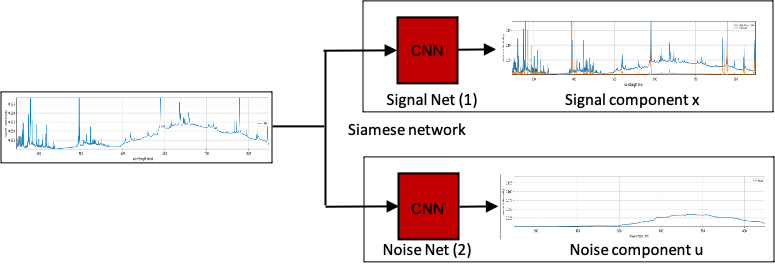}	
	\vspace{-0.75em}
	\caption{SpectroscopyNet CNN Siamese arquitecture.}
	\label{fig:siamese} 
\end{figure}

The deep model utilized is the Siamesse CNN architecture shown in Fig.\ref{fig:siamese} aimed at disentangling in one channel $h_{\Th_1}: \Rspace^N \rightarrow \Rspace^N$ with network weights $\Th_1$ the signal component $\z_{\x}$ and in the second channel  $h_{\Th_2}: \Rspace^N \rightarrow \Rspace^N$ with weights $\Th_2$, noise $\z_n \in \Rspace^N$. To train the Siamesse network we formulate a learning objective 
that randomly draws triplets $\{ \y^j, \y^k, \y^{k'}\}$ with $j \neq k \neq k'$ of unrelated target samples (this requires a weak label to distinguish them) to extract the components $\{ \z^j_{\x}, \z^k_n, \z^{k'}_n\}$ and uses these to minimize the function in Eq. \eqref{loss} as: 
\begin{equation} \label{loss}
\ell = \sum_{j} \left \{ \left \| \y^{j} - (\z^j_{\x} + \z^k_n)  \right \|_{\ell_2} + \langle \z^j_{\x}, \z^k_n\rangle^2 - \sum_{k'}\langle \z^k_n, \z^{k'}_n \rangle \right  \}
\end{equation}
where $\langle \cdot, \cdot \rangle $ denotes an inner product. This, is our modification of the MI InfoNCE objective in Eq. \eqref{MI} that adds an $\ell_2$- norm reconstruction error, jointly trains for the signal channel and constrasts the signal and noise representations, instead.  The training objective of Eq. \eqref{MI} simultaneously promotes: (1) data reconstruction fidelity, (2) consistency reduction between the signal and noise components through a relaxed orthogonality constraint and (3) maximizes alignment or consistency between the noise components. 
The purpose behind this training strategy is to mantain the effects of the noise variable isolated by using representations from unrelated measurements as well as keeping their optimization in a separate channel. This, we hypothesize, effectively blocks and controls under the noise stationarity assumption the information flow between the path $U_N \rightarrow Z_X$ in Fig.\ref{fig:DAG} enabled when conditioning on the measurements $\y$.


%
\section{Experiments} 
\label{Sec:experiments}

The ChemCam laser induced breakdown spectroscopy (LIBS) datasets available at \cite{Washu:2020} were used as the experimentation benchmark for validation of the proposed approach. The datasets contain raw and cleaned spectra obtained from Martian targets (e.g., rocks, soil) and from reference calibration standards of known chemical composition collected on Earth. The specific dataset we employ consists of spectrally resolved LIBS measurements collected on Earth in a laboratory setting from a set of $\sim$ 408 distinct reference calibration standards \cite{Clegg:2017} of known and certified chemical reference composition. 
Each reference standard is repeatedly shot (e.g.,  50 times) following each time measurements of a full 240-905 nm LIBS signal. This process is repeated at multiple locations for each target which amounts to $\sim$41,000 uncleaned LIBS signals. After collection, wavelengths within the bands [240.811,246.635], [338.457,340.797], [382.13,387.859], [473.184,492.427], [849,905.574] were ignored out consistent with practices of the ChemCam team \cite{Clegg:2017}. 

For signal/noise disentanglement we use the Siamesse CNN architecture shown in Fig. \ref{fig:siamese}, each channel as in \cite{castorena2021deep} with $ D = 18$ Conv+BN+ReLu layers optimized through SGD with a learning rate (lr) of 0.1, momentum 0.9 and number of epochs 60. 
At training, the batch sizes consist each of 512 randomly drawn triplets of raw (uncleaned) LIBS measurements with $j \neq k \neq k'$. Each triplet element is randomly chosen (with replacement) from groups of unrelated target measurements.

Qualitative comparisons of the proposed approach are made against the standard method used by the expert practitioners of the ChemCam team to denoise/pre-process the raw LIBS measurements. Quantitative evaluations were performed on the downstream task of calibration. This task relates to the prediction of chemical composition based on the analysis of spectral signatures (e.g., spectral peaks) in the measurements. 
For this task, we test the prediction performance of a neural network (NN) head using a variety of representations as inputs. The compared representations are: (1) raw uncleaned, (2) cleaned as pre-processed by the ChemCam team \cite{Wiens:2013}, and (3) the proposed. Training and testing is performed independently of the representation metdho and using a leave one out split. In other words, multiple linear heads are trained where for each we leave all the data pertaining to one calibration standard out and then use it at test time. This process is repeated multiple times from scratch until all standards are covered, for all of the compared representations and for each element oxide.
%
The head architecture used is a linear layer that takes as input a LIBS measurement of $N=5485$ wavelengths and outputs a scalar $\hat{w}_{\text{oxide}}$ for each oxide $\{ \text{SiO}_2, \text{TiO}, \text{Al}_2\text{O}_3,$ $\text{FeO}_\text{T}, \text{MgO}, \text{CaO}, \text{Na}_2\text{O}, \text{K}_2\text{O}\}$.
\vspace{-0.5em}
\begin{figure}[h]
	\centering 
	\includegraphics[width=0.25\linewidth]{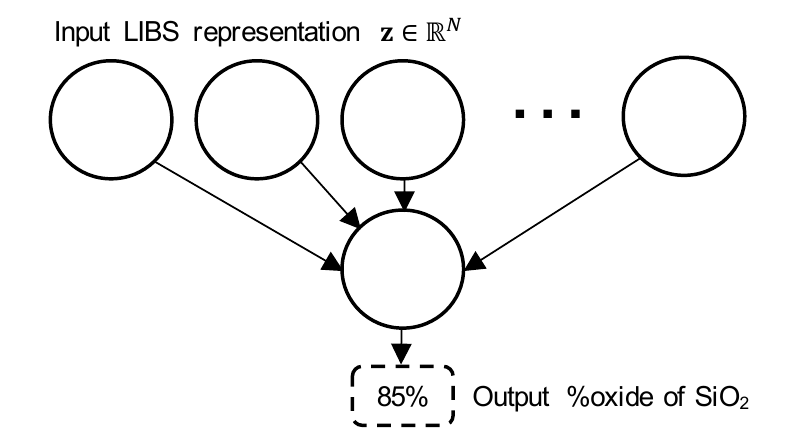}	
	\vspace{-0.75em}
	\caption{Linear head architecture for downstream calibration.}
	\label{fig:head} 
\end{figure}
The hyperparameters used for this linear head are an initial lr of 1.0, decayed after 75 epochs with cosine annealing \cite{loshchilov2016sgdr} and with \#epochs 200. Batches are constructed from a set of $64$ shot-averaged examples randomized over the whole training set without replacement. The shot-averages are computed by averaging the LIBS representations over an individual target and laser shot location. This averaging is consistent with common practices of the ChemCam team \cite{Wiens:2013, Clegg:2017}. 

\subsection{Qualitative Results}
\begin{figure} [h]
	\centering 
	\subfloat[sarm17 (\cite{Wiens:2013}) ]{ \includegraphics[width=0.157\linewidth]{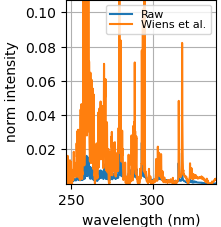} }	
	\subfloat[25tio2 (\cite{Wiens:2013})]{ \includegraphics[width=0.15\linewidth]{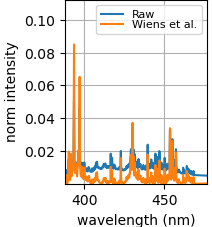} }
	\subfloat[mix1o (\cite{Wiens:2013})]{ \includegraphics[width=0.165\linewidth]{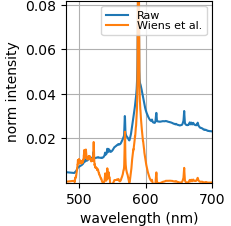} }	
	\subfloat[sarm17  (ours)]{ \includegraphics[width=0.157\linewidth]{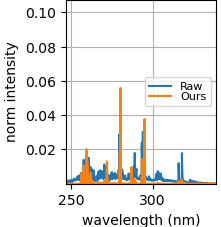} }	
	\subfloat[25tio2 (ours)]{ \includegraphics[width=0.15\linewidth]{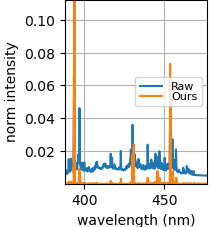} }
	\subfloat[mix1o (ours)]{ \includegraphics[width=0.165\linewidth]{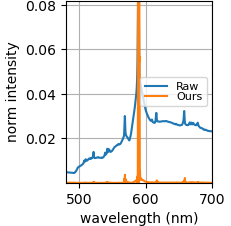} }	
	\caption{Pre-processing example spectrum patches (Targets:sarm17 in (a),(d), 25tio2 in (b),(e) and mix1o in (c),(f), dist:1.6m) in ’Calib’ dataset: raw LIBS signal (blue) pre-processed (orange). Our method yields a representation qualitatively better at removing the noise factor effects.} 
	\label{fig:zoomedpp} 
\end{figure}

A few examples illustrating the qualitative performance of our learned representations compared against uncleaned signals  and cleaned using the hand-crafted feature pre-processing of (Wiens et al.) \cite{Wiens:2013} are included here. Fig.\ref{fig:zoomedpp} shows examples of representative calibration standards spectrum patches with the raw uncleaned measurements shown in blue. Fig.\ref{fig:zoomedpp}.(a-c) of Wiens et al. (in orange), presents spectral lines modulated by what appears to be remains of electron continuum amplified by a system spectral response adjustment. Fig \ref{fig:zoomedpp}.(d-f) compares instead the disentanglement produced by the proposed method (in orange). Note that in all cases the representations in (d-f) seem to cope better in isolating the effects of the target (e.g., peaks) from the noise factors. The spectral content from the non-zero mean dark currents and electron continuum characteristic of the ChemCam instrument \cite{Wiens:2012} are brougth down close to zero better than \cite{Wiens:2013} where remains of both are visually present in all cases.
\begin{figure} [h]
	\centering 
	\subfloat[75tio2 Wiens et al. \cite{Wiens:2013} ]{ \includegraphics[width=1.0\linewidth]{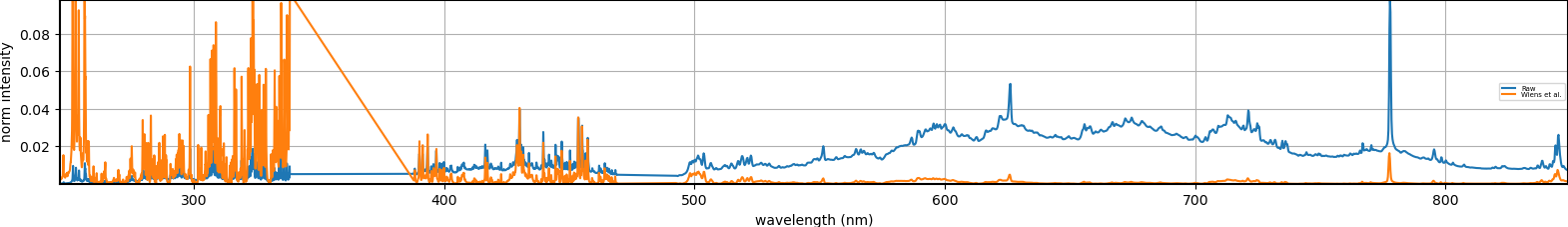} }
	\vspace{-1em}
	\subfloat[75tio2 Ours ]{ \includegraphics[width=1.0\linewidth]{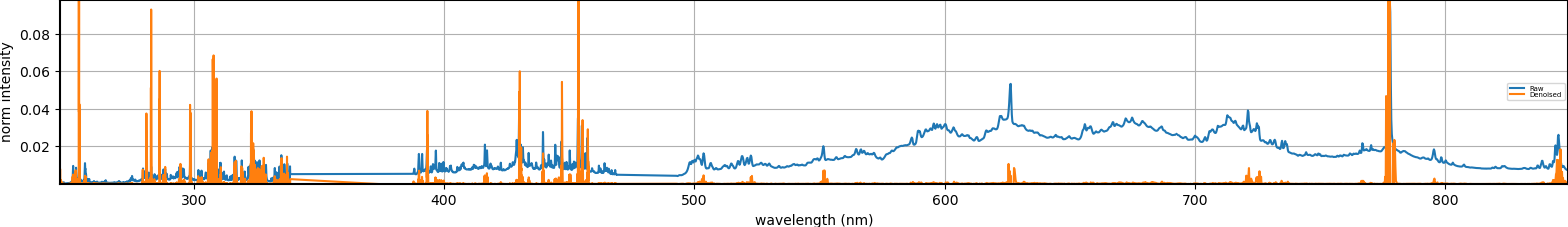} }
	\caption{Pre-processing example (Target:75tio2dist:1.6m): raw LIBS signal in blue, 'cleaned' in orange. Our method is qualitatively better at cleaning the entire UV, VIO, VIS spectrum.} 
	\label{fig:pre-processing} 
\end{figure}
A similar example is shown in Fig.\ref{fig:pre-processing} in this case for the entire LIBS spectrum. We observe that the noise disentanglement properties in Fig. \ref{fig:zoomedpp} are also observed througout the entire ultraviolet (UV), visible (VIO) and near infrared (VIS) spectrum.

\begin{figure*} [t]
	\captionsetup[subfigure]{labelformat=empty}
	\centering 
	\subfloat{ \includegraphics[width=0.15\linewidth]{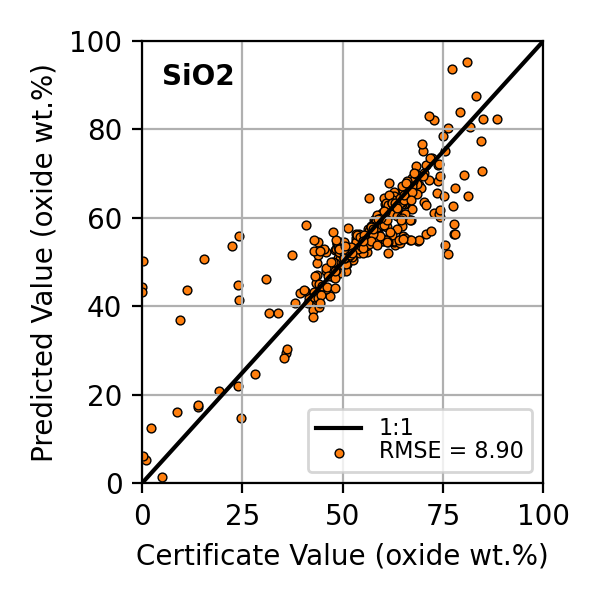} }
	\subfloat{ \includegraphics[width=0.15\linewidth]{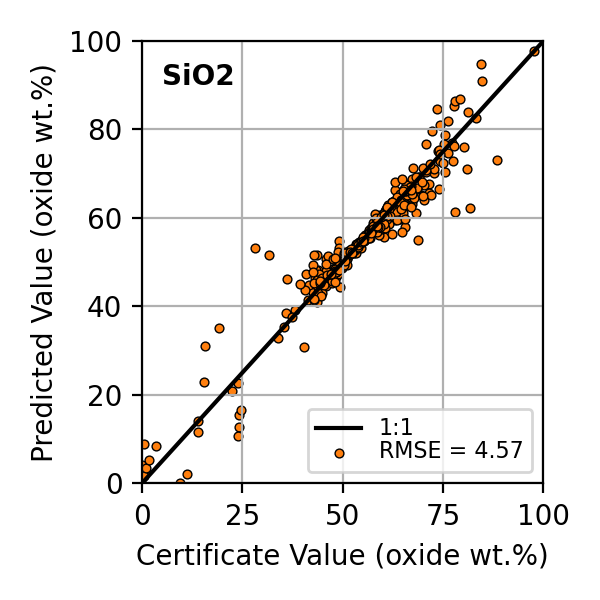} }
	\subfloat{ \includegraphics[width=0.15\linewidth]{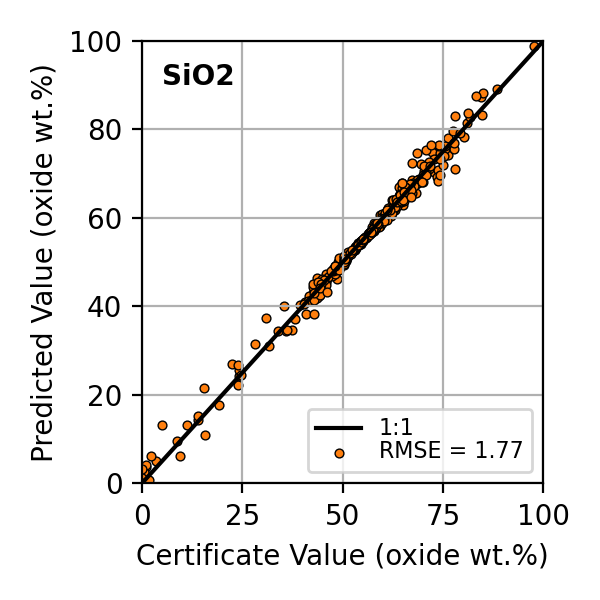} }
	\subfloat{ \includegraphics[width=0.15\linewidth]{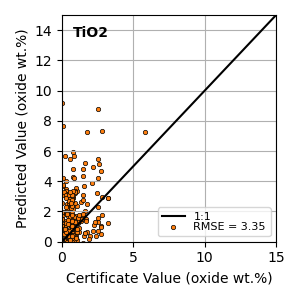} }
	\subfloat{ \includegraphics[width=0.15\linewidth]{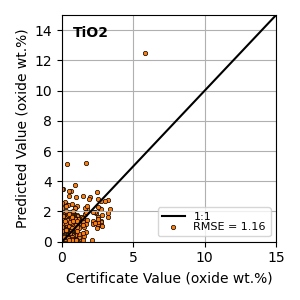} }
	\subfloat{ \includegraphics[width=0.15\linewidth]{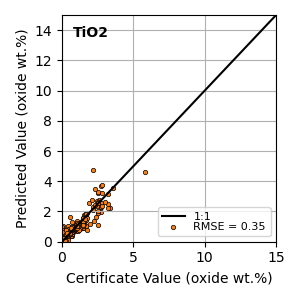} }
	\vspace{-1em}
	
	\subfloat{ \includegraphics[width=0.15\linewidth]{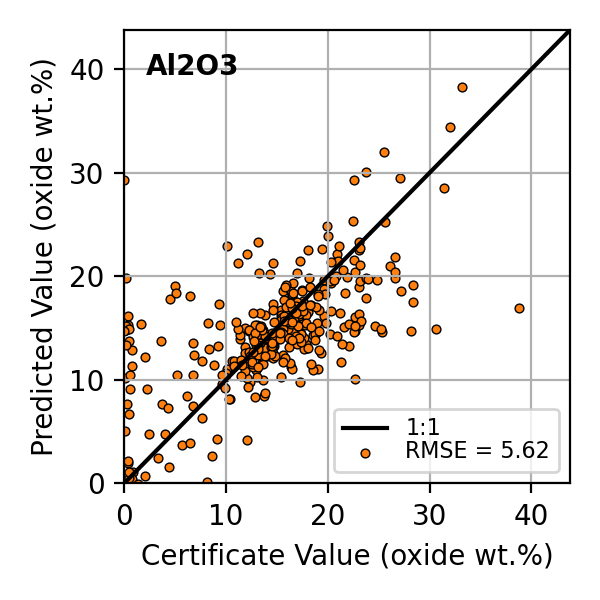} }
	\subfloat{ \includegraphics[width=0.15\linewidth]{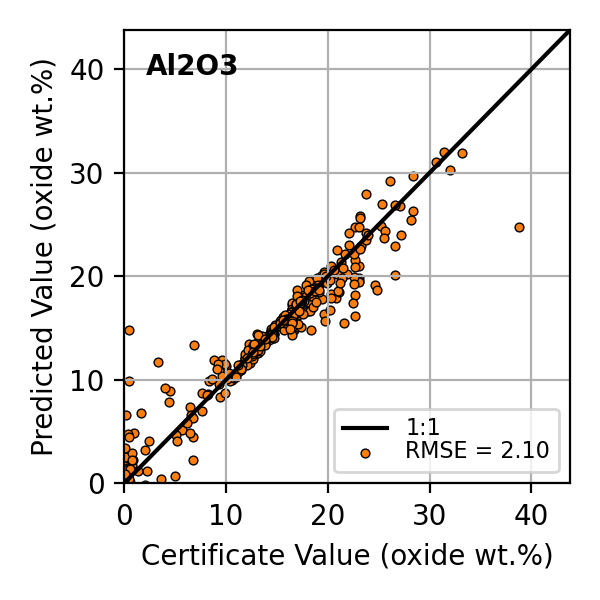} }
	\subfloat{ \includegraphics[width=0.15\linewidth]{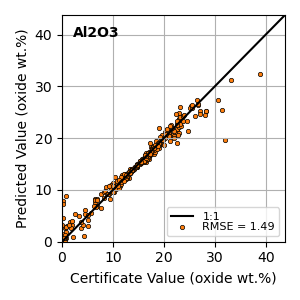} }
	\subfloat{ \includegraphics[width=0.15\linewidth]{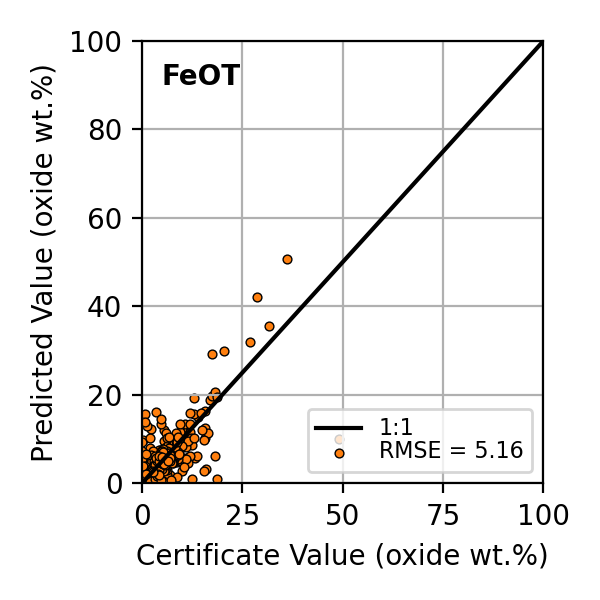} }
	\subfloat{ \includegraphics[width=0.15\linewidth]{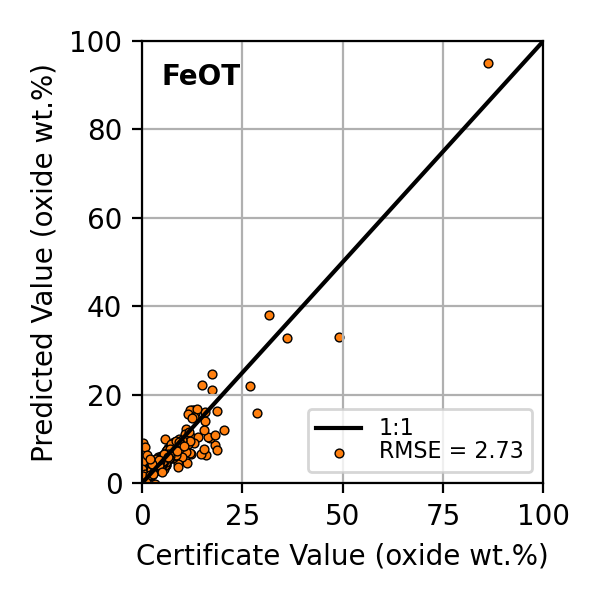} }
	\subfloat{ \includegraphics[width=0.15\linewidth]{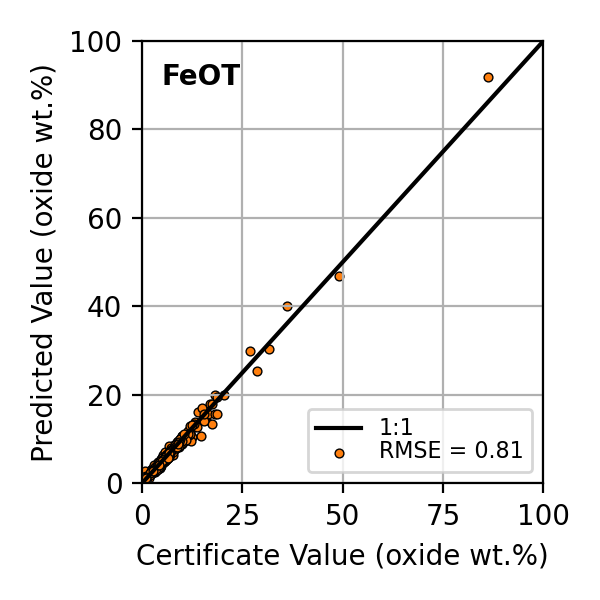} }
	\vspace{-1em}
	
	\subfloat{ \includegraphics[width=0.15\linewidth]{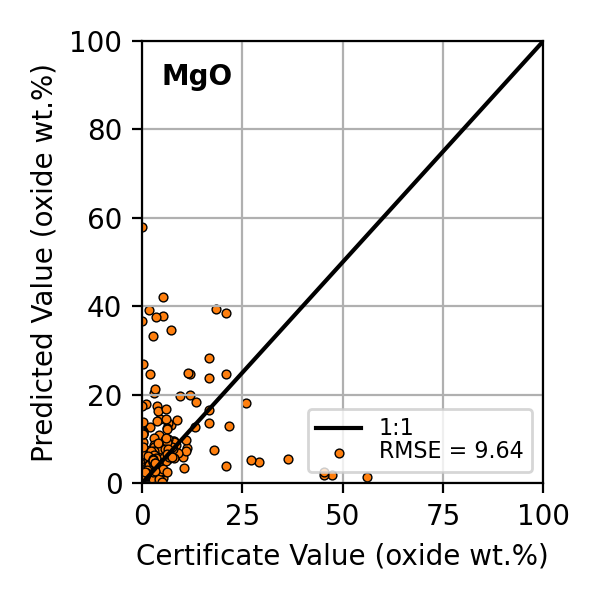} }
	\subfloat{ \includegraphics[width=0.15\linewidth]{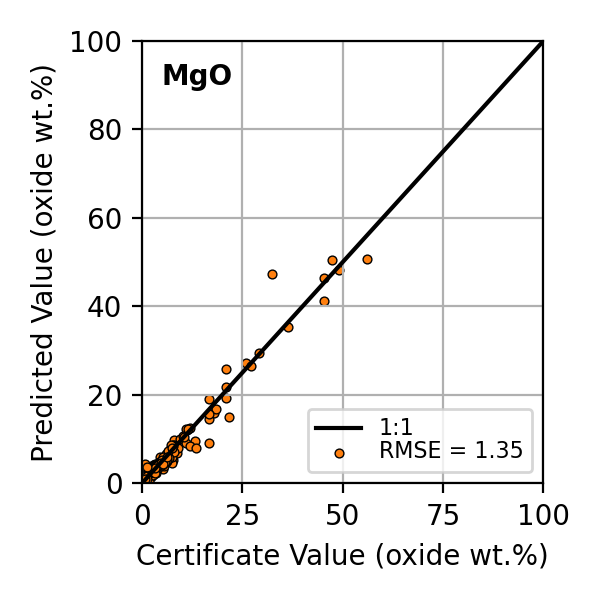} }
	\subfloat{ \includegraphics[width=0.15\linewidth]{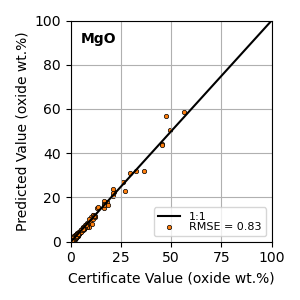} }
	\subfloat{ \includegraphics[width=0.15\linewidth]{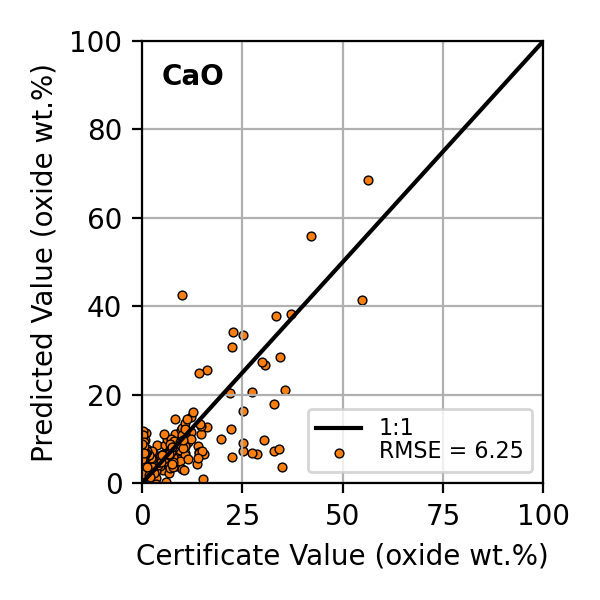} }
	\subfloat{ \includegraphics[width=0.15\linewidth]{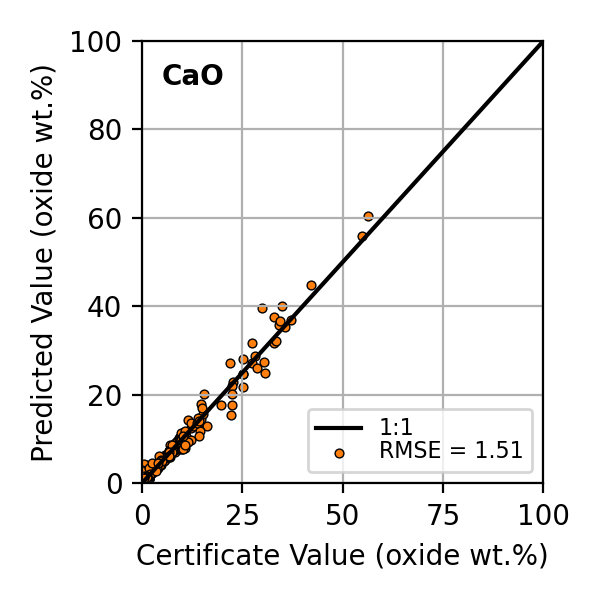} }
	\subfloat{ \includegraphics[width=0.15\linewidth]{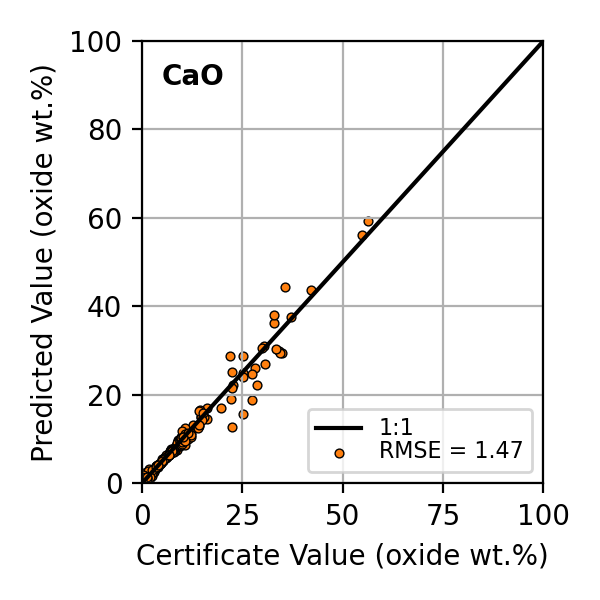} }
	\vspace{-1em}
	
	\subfloat[(a) Raw LIBS signal ]{ \includegraphics[width=0.15\linewidth]{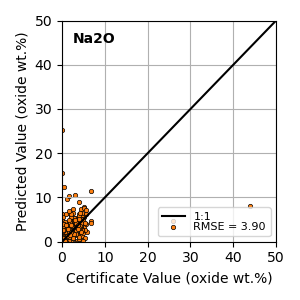} }
	\subfloat[(b) Wiens et al. \cite{Wiens:2013} ]{ \includegraphics[width=0.15\linewidth]{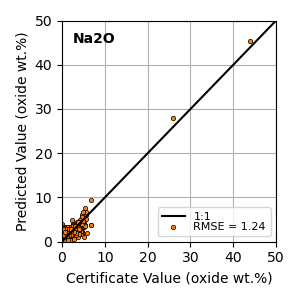} }
	\subfloat[(c) Ours]{ \includegraphics[width=0.15\linewidth]{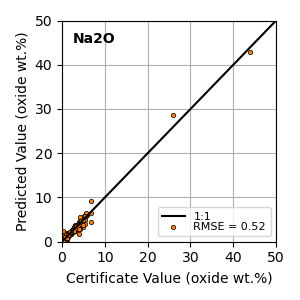} }
	\subfloat[(d) Raw LIBS signal ] { \includegraphics[width=0.15\linewidth]{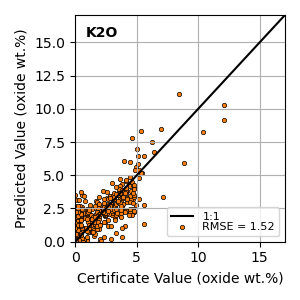} }
	\subfloat[(e) Wiens et al. \cite{Wiens:2013} ]{ \includegraphics[width=0.15\linewidth]{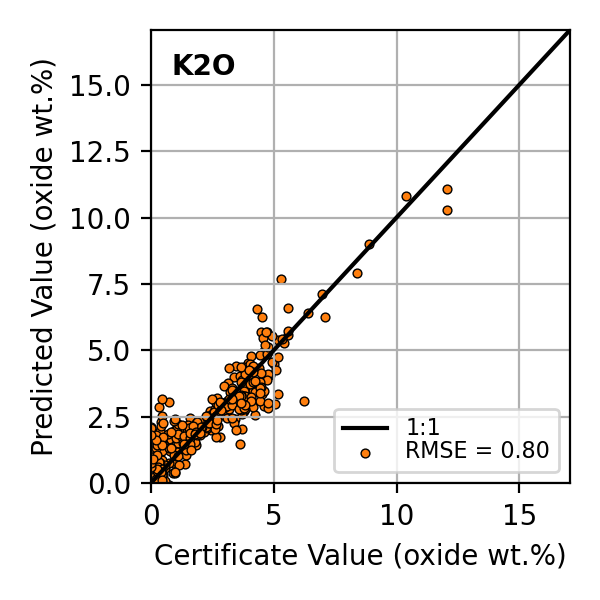} }
	\subfloat[(f) Ours]{ \includegraphics[width=0.15 \linewidth]{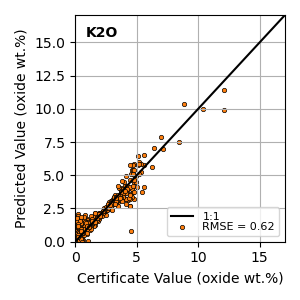} }
	
	\caption{Disentanglement effectivements in chemical composition predictions. The proposed method resulted in disentanglements that lead to less spread and lower RMSE predictions from the ideal 1:1 line and compared to the raw and the ChemCam team's pre-processing in \cite{Wiens:2013} representations. Included are the regression results of the 8 major chemical elements in the $\sim 470$ reference standards 'Calib' dataset. Corresponding RMSE's for each element are shown in the legend for each oxide.}
	\label{fig:comparison} 
\end{figure*}
\begin{figure} [h]
	\centering 
	\subfloat[RMSE.]{ \includegraphics[width=0.45\linewidth]{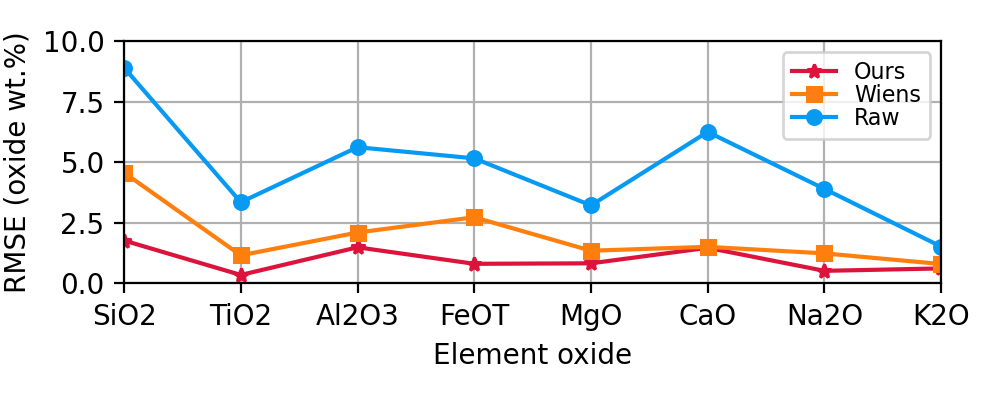} }
	\subfloat[MAXE.]{ \includegraphics[width=0.45\linewidth]{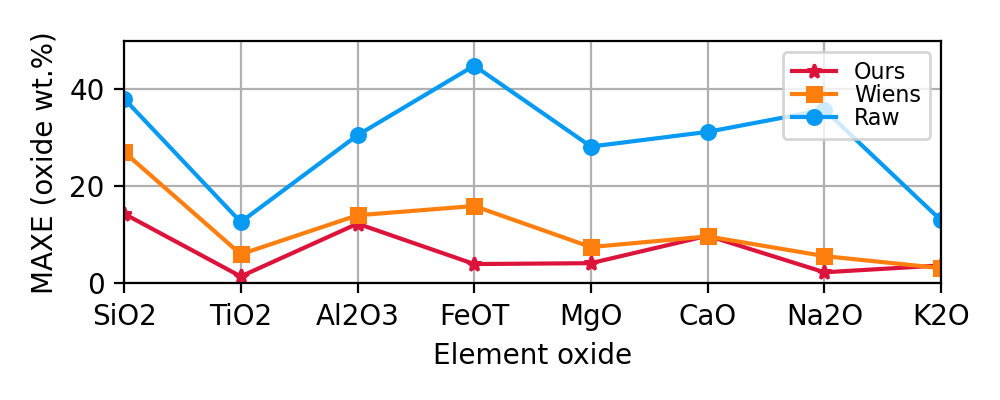} }
	\caption{ Performance as a function of chemical oxide.} 
	\label{fig:performance} 
\end{figure}

\subsection{Quantitative Results.}
Quantitative performance results are included as plots of prediction (in y-axis) versus ground truth $\%$-oxide (in x-axis) for each of the eight major elements (labeled inside each plot). 
Fig.\ref{fig:comparison} columns (a,d), (b,e) and (c,f) are the results of the raw, pre-processing of \cite{Wiens:2013} and the proposed learned representations, respectively. 
Here, we observe that for all three representations and element oxides the predictions (shown as orange dots) follow in some way or another the ideal 1:1 line in black. The raw representation leads to a higher spread of calibration predictions over the 1:1 line, followed by the pre-procesing of \cite{Wiens:2013} while the proposed leads to a relatively better and smaller spread. This is quantified by means of the root mean squared error (RMSE) shown in the legend of each subfigure where the proposed learned representation outperforms all others throughout all element oxide cases.

The reduced spread and RMSE resulting from the proposed framework in the downstream prediction task demonstrate that our method is effective in isolating the effects of noise from the target while keeping the meaningfull target features, at least in what regards to the eight major element oxides. Moreover, it was able to outperform in a calibration task the current hand-crafted ChemCam's team method of choice for denoising. Compared to the raw uncleaned representation where no target information has been lost from the application ofdisentanglement, we observe that our method is able to mantain the effects of the target without losing any meaningful information. 
With this, we thus corroborate that the isolation of noise/target effects in LIBS is possible in a weakly-supervised way using deep models and our learning formulation described in Eq. \eqref{loss}. This, without assuming any explicit priors or statistical characterizations of the noise factors other than stationarity and by leveraging the vast amount of data redundancies that exist in LIBS data. In terms of computational complexity, we would like to note that our proposed method can generate cleaned representations at $\sim$50 Hz on a CPU when in feed-forward mode similar to \cite{castorena2021deep}. This renders our method more computationally efficient than those described in \cite{Wiens:2013, Clegg:2017}.


\section{Conclusion}
\label{Sec:conclusion} 
In this work, we proposed an approach to learn to clean spectroscopy signals without supervision on clean data. Our formulation imposes identifiability constraints to disentangle the effects of noise from target by leveraging noise stationarity across measurements and without assuming a known centered noise distribution. The proposed deep Siamese model uses triplets of unrelated measurements aimed at simultaneously extracting the mutual information between them for noise disentanglement while reducing consistency with the signal component in an $\ell_2$-norm reconstruction fidelity constraint. Our experimentation on laser induced breakdown spectroscopy (LIBS) data shows that the learned disentanglements are qualitatively and quantitatively more effective and efficient than the hand-crafted pre-processing methods currently used by the ChemCam team.

\section*{Acknowledgement}
Research was supported by the Laboratory Directed Research and Development program of Los Alamos National Laboratory under project number LDRD-20210043DR.


\bibliographystyle{elsarticle-num} 

\end{document}